\documentclass{pbmlarxiv}
\usepackage{float}
\usepackage{xcolor}
\usepackage{multirow}
\usepackage{booktabs}
\usepackage{wrapfig}
\definecolor{darkgreen}{RGB}{0,100,0}
\newcommand*{\regularbox}[1]{\color{black}
	\setlength{\fboxsep}{-1\fboxrule}
	\fbox{\hspace{1.1pt}\textbf{\strut#1}\hspace{1.2pt}}
	\color{black}  }
\usepackage[nameinlink,capitalize,noabbrev]{cleveref}

\begin{document}

	\title{NMT-Keras: a Very Flexible Toolkit with a Focus\titlelinebreak{}{} on Interactive NMT and Online Learning}

	\institute{prhlt}{Pattern Recognition and Human Language Technology Research Center, Universitat Polit\`ecnica de Val\`encia, Spain}

	\author{
		firstname=Álvaro,
		surname=Peris,
		institute=prhlt,
		corresponding=yes,
		email={lvapeab@prhlt.upv.es},
		address={Pattern Recognition and Human Language Technology Research Center,\\ Universitat  Politècnica de València, \\
			Camino de Vera s/n, 46022 Valencia, SPAIN.}
	}

	\author{
		firstname=Francisco,
		surname=Casacuberta,
		institute=prhlt,
		corresponding=no,
		email={fcn@prhlt.upv.es},
		address={Pattern Recognition and Human Language Technology Research Center,\\ Universitat Politècnica de València, \\
			Camino de Vera s/n, 46022 Valencia, SPAIN.}
	}

	\shorttitle{NMT-Keras}
	\shortauthor{Á. Peris and F. Casacuberta}

	\PBMLmaketitle

	\begin{abstract}
		We present NMT-Keras, a flexible toolkit for training deep learning models, which puts a particular emphasis on the development of advanced applications of neural machine translation systems, such as interactive-predictive translation protocols and long-term adaptation of the translation system via continuous learning.
		NMT-Keras is based on an extended version of the popular Keras library, and it runs on Theano and TensorFlow. State-of-the-art neural machine translation models are deployed and used following the high-level framework provided by Keras.
		Given its high modularity and flexibility, it also has been extended to tackle different problems, such as image and video captioning, sentence classification and visual question answering.
	\end{abstract}

	\section{Introduction}
	\label{sec:intro}
	To easily develop new deep learning models is a key feature in this fast-moving field. We introduce  NMT-Keras\footnote{\url{https://github.com/lvapeab/nmt-keras}}, a flexible toolkit for neural machine translation (NMT), based on the Keras library for deep learning \citep{Chollet15}. Keras is an API written in Python which provides high-level interfaces to numerical computation libraries, such as Theano \citep{Theano16} or TensorFlow \citep{Abadi16}. Keras is well-structured and documented, with designing principles that make it modular and extensible, being easy to construct complex models.

	Following the spirit of the Keras library, we developed NMT-Keras, released under MIT license, that aims to provide a highly-modular and extensible framework to NMT. NMT-Keras supports advanced features, including support of interactive-predictive NMT (INMT) \citep{Barrachina09,Peris17a} protocols, continuous adaptation \citep{Peris17b} and active learning \citep{Peris18b} strategies. An additional goal, is to ease the usage of the library, but allowing the user to configure most of the options involving the NMT process.

	Since its introduction \citep{Sutskever14,Cho14}, NMT has advanced by leaps and bounds. Several toolkits currently offer fully-fledged NMT systems. Among them, we can highlight OpenNMT \citep{Klein17}, Tensor2Tensor \citep{Vaswani18} or Nematus \citep{Sennrich17}. NMT-Keras differentiates from them by offering interactive-predictive and long-term learning functionalities, with the final goal of fostering a more productive usage of the NMT system.

	This document describes the main design and features of NMT-Keras. First, we review the deep learning framework which is the basis of the toolkit. Next, we summarize the principal features and functionality of the library. We conclude by showing translation and INMT results.

	\section{Design}

	NMT-Keras relies on two main libraries: a modified version of Keras\footnote{\url{https://github.com/MarcBS/keras}}, which provides the framework for training the neural models; and a wrapper around it, named Multimodal Keras Wrapper\footnote{\url{https://github.com/lvapeab/multimodal_keras_wrapper}}, designed to ease the usage of Keras and the management of data.These tools represent a general deep learning framework, able to tackle different problems, involving several data modalities. The problem of NMT is an instantiation of the sequence-to-sequence task, applied to textual inputs and outputs. NMT-Keras is designed to tackle this particular task. The reason for relying on a fork of Keras is because this allows us to independently design functions for our problems at hand, which may be confusing for the general audience of Keras. However, in a near future we hope to integrate our contributions into the main Keras repository. In this section, we describe these components and their relationships in our framework.

	\subsection{Keras}

	As mentioned in \cref{sec:intro}, Keras is a high-level deep learning API, which provides a neat interface to numerical computation libraries. Keras  allows to easily implement complex deep learning models by defining the layers as building blocks. This simplicity, together with the quality of its code, has made Keras to be one of the most popular deep learning frameworks. It is also well-documented, and supported by a large community, which fosters its usage.

	In Keras, a model is defined as a directed graph of layers or operations, containing one or more inputs and one or more outputs. Once a model has been defined, it is compiled for training, aiming to minimize a loss function. The optimization process is carried out, via stochastic gradient descent (SGD), by means of an optimizer.

	Once the model is compiled, we feed it with data, training it as desired. Once a model is trained, it is ready to be applied on new input data.

	\subsection{Multimodal Keras Wrapper}
	The Multimodal Keras Wrapper allows to handle the training and application of complex Keras models, data management (including multimodal data) and application of additional callbacks during training. The wrapper defines two main objects and includes a number of extra features:

	\begin{description}
		\item [Dataset:] A Dataset object is a database adapted for Keras, which acts as data provider. It manages the data splits (training, validation, test). It accepts several data types, including text, images, videos and categorical labels. In the case of text, the Dataset object builds the vocabularies, loads and encodes text into a numerical representation and also decodes the outputs of the network into natural language. In the case of images or videos, it also normalizes and equalizes the images; and can apply data augmentation.

		\item [Model wrapper:] This is the core of the wrapper around Keras. It connects the Keras  library with the Dataset object and manages the functions for training and applying the Keras models. When dealing with sequence-to-sequence models, it implements a beam search procedure. It also contains a training visualization module and ready-to-use convolutional neural networks (CNN) architectures.

		\item [Extra:] Additional functionalities include extra callbacks, I/O management and evaluation of the system outputs. For computing the translation quality metrics of the models, we use the coco-caption evaluation tool \citep{Chen15}, which provides common evaluation metrics: BLEU, METEOR, CIDEr, and ROUGE-L. Moreover, we modified it\footnote{\url{https://github.com/lvapeab/coco-caption}} to also include TER.
	\end{description}

	\subsection{NMT-Keras}
	The NMT-Keras library makes use of the aforementioned libraries, for building a complete NMT toolkit. The library is compatible with Python $2$ and $3$. The training of NMT models is done with the \texttt{main.py} file. The hyperparameters are set via a configuration file (\texttt{config.py}), and can also be set from the command line interface. To train a NMT system with NMT-Keras is straightforward:
	\begin{enumerate}
		\item Set the desired configuration in \texttt{config.py}.
		\item Launch \texttt{main.py}.
	\end{enumerate}

	The training process will then prepare the data, constructing the Dataset object and the Keras model. The default models implemented in NMT-Keras are an attentional RNN encoder--decoder \citep{Bahdanau15,Sennrich17}, and the Transformer model \citep{Vaswani17}. Once the model is compiled, the training process starts, following the specified configuration. For translating new text with a trained model, we use beam search.

	\section{Features}

	As we keep our Keras fork constantly up-to-date with the original library, NMT-Keras has access to the full Keras functionality, including (but not limited to):

	\begin{description}
		\item [Layers:] Fully-connected layers, CNN, recurrent neural networks (RNN), including long short-term memory (LSTM) \citep{Hochreiter97}, gated recurrent units (GRU) \citep{Cho14} and their bidirectional \citep{Schuster97} version, embeddings, noise, dropout \citep{Srivastava14} and batch normalization \citep{Ioffe15} layers.
		\item [Initializers:] The weights of a model can be initialized to a constant value, to values drawn from statistical distributions or according to the strategies proposed by \citet{Glorot10}, \citet{He15} and \citet{Klambauer17}.
		\item [Optimizers:] \label{sec:optimizers}A number of SGD variants are implemented: vanilla SGD, RMSProp \citep{Tieleman12}, Adagrad \citep{Duchi11}, Adadelta \citep{Zeiler12}, Adam and Adamax \citep{Kingma14}. The learning rate can be scheduled according to several strategies (linear, exponential, ``noam" \citep{Vaswani17}).
		\item [Regularizers and constraints:] Keras allows to set penalties and constraints to the parameters and to the layer outputs of the model.
	\end{description}

	Our version of Keras implements additional layers, useful for sequence-to-sequence problems:
	\begin{description}
		\item [Improved RNNs:] All RNN architectures can be used in an autoregressive way, i.e. taking into account the previously generated token. They also integrate attention mechanisms, supporting the \textit{add} \citep{Bahdanau15} and \textit{dot} \citep{Luong15a} models.
		\item [Conditional RNNs:] Conditional \citep{Sennrich17} LSTM/GRU layers, consisting in cascaded applications of LSTM/GRU cells, with attention models in between.
		\item [Multi-input RNNs:] LSTM/GRU networks with two and three different inputs and independent attention mechanisms \citep{Bolanos18}.
		\item [Transformer layers:] Multi-head attention layers, positional encodings and position-wise feed-forward networks \citep{Vaswani17}.
		\item [Convolutional layers:] Class activation maps \citep{Zhou16}.
	\end{description}

	Finally, NMT-Keras supports a number of additional options. Here we list the most important ones, but we refer the reader to the library documentation\footnote{\url{https://nmt-keras.readthedocs.io}} for an exhaustive listing of all available options:
	\begin{description}
		\item [Deep models:] Deep residual RNN layers, deep output layers \citep{Pascanu14} and deep fully-connected layers for initializing the state of the RNN decoder.
		\item [Embeddings:] Incorporation of pre-trained embeddings in the NMT model and embedding scaling options.
		\item [Regularization strategies:] Label smoothing \citep{Szegedy15}, early-stop, weight decay, doubly stochastic attention regularizer \citep{Xu15} and a fine-grained application of dropout.
		\item [Search options:] Normalization of the beam search process by length and coverage penalty. The search can be also constrained according to the length of the input/output sequences.
		\item [Unknown word replacement:] Replace unknown words according to the attention model \citep{Jean15}. The replacement may rely on a statistical dictionary.
		\item [Tokenizing options:] Including full support to byte-pair-encoding  \citep{Sennrich16}.
		\item [Integration with other tools:] Support for Spearmint \citep{Gelbart14}, for Bayesian optimization of the hyperparameters and Tensorboard, the visualization tool of TensorFlow.\\
	\end{description}

	Apart from these model options, NMT-Keras also contains scripts for ensemble decoding and generation of $N$-best lists; sentence scoring, model averaging and construction of statistical dictionaries (for unknown words replacement). It also contains a client-server architecture, which allows to access to NMT-Keras via web. Finally, in order to introduce newcomers to NMT-Keras, a set of tutorials are available, explaining step-by-step how to construct a NMT system.

	\subsection{Interactive-predictive machine translation}
	\label{sec:INMT}

	The interactive-predictive machine translation (IMT) framework constitutes an efficient alternative to the regular post-editing of machine translation; with the aim of obtaining high-quality translations minimizing the human effort required for this \citep{Barrachina09}. IMT is a collaborative symbiosis between human and system, consisting in an iterative process in which, at each iteration, the user introduces a correction to the system hypothesis. The system takes into account the correction and provides an alternative hypothesis, considering the feedback from the user. \cref{fig:offline-INMT} shows an example of the IMT protocol.

	\begin{wrapfigure}{r}{0.5\textwidth}
		\centering
		\def\arraystretch{1.3}
		\footnotesize
		\begin{tabular}{ccl}
			\toprule
			\multicolumn{2}{l}{Source:}& They are lost forever . \\
			\multicolumn{2}{l}{Target:}& Ils sont perdus à jamais . \\
			\hline
			\textbf{0} & \textbf{MT} & Ils sont perdus pour toujours . \\
			\midrule
			\multirow{2}{*}{\textbf{1}} & $\bf{User}$& \textcolor{darkgreen}{\textit{Ils sont perdus}} \regularbox{à} pour toujours .\\
			& $\bf{MT}$ &\textcolor{darkgreen}{ \textit{Ils sont perdus à}} jamais . \\
			\midrule
			\textbf{2} & \textbf{User} &\textcolor{darkgreen}{\textit{Ils sont perdus à jamais .}} \\
			\bottomrule
		\end{tabular}

		\caption{\label{fig:offline-INMT} IMT session. The user introduces in iteration 1 a character correction (boxed). The MT system modifies its hypothesis, taking into account this feedback.
		}
	\end{wrapfigure}

\begin{figure}[h]
		\centering
		\includegraphics[width=0.95\linewidth]{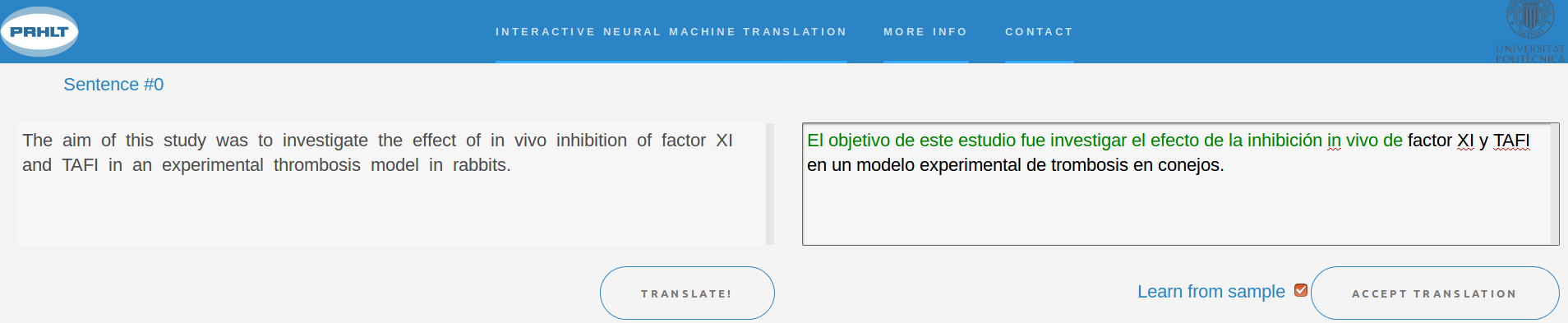}
		\caption{\label{fig:frontend}Frontend of the INMT with OL website built with NMT-Keras. In the left-side, the source sentence to translate is displayed. The system computes a translation hypothesis, located at the right frame. The user makes changes on this hypothesis. As a key is pressed, the hypothesis is immediately updated, taking into account the user correction. In this interaction protocol, the system validates the hypothesis prefix (validated in green), although we expect to support more flexible protocols \citep{Peris17a} in a future. This process is repeated until the desired translation is reached. Then, the user presses the \textit{Accept translation} button, validating the translation. The system will use this sample to update the model if the ``\textit{Learn from sample}'' checkbox is activated. The demo is available at: \url{http://casmacat.prhlt.upv.es/inmt}.}
	\end{figure}

	This protocol has show to be especially effective when combined with NMT \citep{Knowles16,Peris17a}. NMT-Keras implements the interactive protocols described by \citet{Peris17a}. Moreover, they can be combined with online learning (OL) or active learning methods, which allow to specialize the system into a given domain or to the preferences of a user \citep{Peris18}. These protocols are also implemented in NMT-Keras. We built a demo website\footnote{\url{http://casmacat.prhlt.upv.es/inmt}.} of these interactive, adaptive systems using the client-server features of the toolkit (\cref{fig:frontend}). 

	\subsection{Tackling other tasks}

	The modular design of Keras and Multimodal Keras Wrapper allows to use them to tackle other problems, following the spirit of NMT-Keras. These libraries have been used to perform video captioning \citep{Peris16}, egocentric video captioning considering temporal contexts \citep{Bolanos18}, text classification \citep{Peris17d}, food recognition and localization \citep{Bolanos16,Bolanos17c} and visual question answering \citep{Bolanos17}.

\section{Results}

We report now results of NMT-Keras, assessing translation quality and INMT effectiveness. Due to space restrictions, we report results on two tasks: EU \citep{Barrachina09} and Europarl \citep{Koehn05}. More extensive results obtained with NMT-Keras can be found at \citet{Peris18}. For the first task, we used the standard partitions. For the Europarl corpus, we used the \texttt{newstest2012} and \texttt{newstest2013}, as development and test, in order to establish a comparison with other IMT works \citep[e.g.][]{Ortiz16}.
The NMT system was configured as in \citet{Peris18}. For the sake of comparison, we include results of phrase-based statistical machine translation (PB-SMT), using the standard setup of \texttt{Moses} \citep{Koehn07}. We computed significance tests ($95\%$) via approximate randomization and confidence intervals with bootstrap resampling \citep{Riezler05}.

\cref{table:results-mt-quality} shows the translation quality. NMT-Keras outperformed \texttt{Moses}, obtaining significant TER and BLEU improvements almost every language pair. Only in one case \texttt{Moses} obtained better TER than NMT-Keras.

\begin{table}[!h]
	\centering
	\begin{tabular}{llllll}
		\toprule
		& & \multicolumn{2}{c}{TER ($\downarrow$)}
		& \multicolumn{2}{c}{ BLEU ($\uparrow$)}
		\\ \cmidrule(lr){3-4}\cmidrule(lr){5-6}
		& &PB-SMT& NMT& PB-SMT& NMT\\
		\midrule
		\multicolumn{1}{ c }{\multirow{2}{*}{EU} } &
		En$\rightarrow$De   & $54.1 \pm 1.9$ & $\mathbf{52.3 \pm 1.9}$  & $35.4 \pm 2.1$ & $\mathbf{36.4 \pm 2.0}$  \\
		& En$\rightarrow$Fr   & $41.4 \pm 1.6$ & $\mathbf{38.4 \pm 1.5}$ & $47.8\pm 1.7$ & $\mathbf{50.4 \pm 1.6}$  \\
		\midrule
		\multicolumn{1}{ c }{\multirow{2}{*}{Europarl} } &

		En$\rightarrow$De  & $\mathbf{62.2 \pm 0.3}$ & $63.1 \pm 0.4$ & $ 19.5 \pm 0.3$ & $\mathbf{20.0 \pm 0.3}$ \\
		\multicolumn{1}{ c  }{}                        &
		En$\rightarrow$Fr  & $56.1 \pm 0.3$ & $\mathbf{55.0 \pm 0.3}$ & $26.5 \pm 0.3$ & $\mathbf{27.8 \pm 0.3}$  \\
		\bottomrule
	\end{tabular}
	\caption{\label{table:results-mt-quality}Results of translation quality for all tasks in terms of TER [\%] and BLEU [\%]. Results that are statistically significantly better for each task and metric are boldfaced. Extended results can be found in \citet{Peris18}.}
\end{table}

\subsection{Interactive NMT}

	We evaluate now  the performance of NMT-Keras on an IMT scenario. We are interested in measuring the effort that the user must spend in order to achieve the desired translation. Due to the prohibitive cost that an experimentation with real users conveys, the users were simulated. The references of our datasets were considered to be the desired translations. The amount of effort was measured according to keystroke mouse-action ratio (KSMR) \citep{Barrachina09}, defined as the number of keystrokes plus the number of mouse-actions required for obtaining the desired sentence, divided by the number of characters of such sentence.

\begin{table}[!h]
	\centering

	\begin{tabular}{llrr}
		\toprule
		& & \multicolumn{2}{c}{KSMR [\%] ($\downarrow$)}
		\\ \cmidrule(lr){3-4}
		& &INMT & SOTA \\
		\midrule
		\multicolumn{1}{ c }{\multirow{2}{*}{EU} } &
		En$\rightarrow$De & $\mathbf{19.8 \pm 0.5}$ &   $30.5 \pm 1.1$$^\ddagger$ \\
		\multicolumn{1}{ c  }{}                        &
		En$\rightarrow$Fr   & $\mathbf{16.3 \pm 0.4}$ & $ 25.5 \pm 1.1$$^\ddagger$\\
		\midrule
		\multicolumn{1}{ c }{\multirow{2}{*}{Europarl} } &
		En$\rightarrow$De    & $\mathbf{32.9 \pm 0.2}$ & $ 49.2 \pm  0.4$$^\dagger$\\
		\multicolumn{1}{ c  }{}                        &
		En$\rightarrow$Fr & $\mathbf{29.8 \pm 0.2}$ & $44.4 \pm 0.5 $$^\dagger$ \\
		\bottomrule
	\end{tabular}
	\caption{\label{table:results-task1-literature} Effort required by INMT systems compared to the state-of-the-art (SOTA), in terms of KSMR [\%]. Results that are statistically significantly better for each task and metric are boldfaced. $^\dagger$ refers to \citet{Ortiz16}, $^\ddagger$ to \citet{Barrachina09}. Extended results can be found in \citet{Peris18}.}
\end{table}

	\cref{table:results-task1-literature} shows the performance in KSMR (\%) of the INMT systems. We also compare these results with the best results obtained in the literature for each task. All INMT systems outperformed by large the other ones. Again, we refer the reader to \citet{Peris18} for a larger set of experiments, including OL.

	\section{Conclusions}

	We introduced NMT-Keras, a toolkit built on the top of Keras, that aims to ease the deployment of complex NMT systems by having a modular and extensible design. NMT-Keras has a strong focus on building adaptive and interactive NMT systems; which leverage the effort reduction of a user willing to obtain high-quality translations. Finally, its flexibility allows the NMT-Keras framework to be applied directly to other problems, such as multimedia captioning or sentence classification.

	We intend to continue the active development of the tool, including new functionalities and improving the quality of the source code. Moreover, we hope to integrate our tool into the Keras ecosystem in a near future.

	\section*{Acknowledgements}

	Much of our Keras fork and the Multimodal Keras Wrapper libraries were developed together with Marc Bolaños. We also acknowledge the rest of contributors to these open-source projects. The research leading this work received funding from grants PROMETEO/2018/004 and CoMUN-HaT - TIN2015-70924-C2-1-R. We finally acknowledge NVIDIA Corporation for the donation of GPUs used in this work.

	\bibliography{mybib}

	\correspondingaddress
\end{document}